\documentclass{article}

\usepackage{microtype}
\usepackage{graphicx}
\usepackage{subfigure}
\usepackage{booktabs} 

\usepackage{hyperref}

\usepackage[accepted]{icml2025}

\usepackage{amsmath}
\usepackage{amssymb}
\usepackage{mathtools}
\usepackage{amsthm}

\usepackage[capitalize,noabbrev]{cleveref}

\theoremstyle{plain}

\theoremstyle{definition}

\theoremstyle{remark}

\usepackage[textsize=tiny]{todonotes}

\icmltitlerunning{Accelerated 
co-design of robots
through morphological pretraining}

\begin{document}

\twocolumn[
\icmltitle{Accelerated 
co-design of robots
through morphological pretraining}



\begin{icmlauthorlist}
\icmlauthor{Luke Strgar}{n}
\icmlauthor{Sam Kriegman}{n}
\end{icmlauthorlist}

\icmlaffiliation{n}{Northwestern University, Evanston, IL, USA}

\icmlcorrespondingauthor{Luke Strgar}{lvs@u.northwestern.edu}
\icmlcorrespondingauthor{Sam Kriegman}{sam.kriegman@northwestern.edu}

\icmlkeywords{Phototaxis, Soft robot}

\vskip 0.3in
]

\printAffiliationsAndNotice{}

\begin{figure*}[t]
    \centering
    \includegraphics[width=\textwidth]{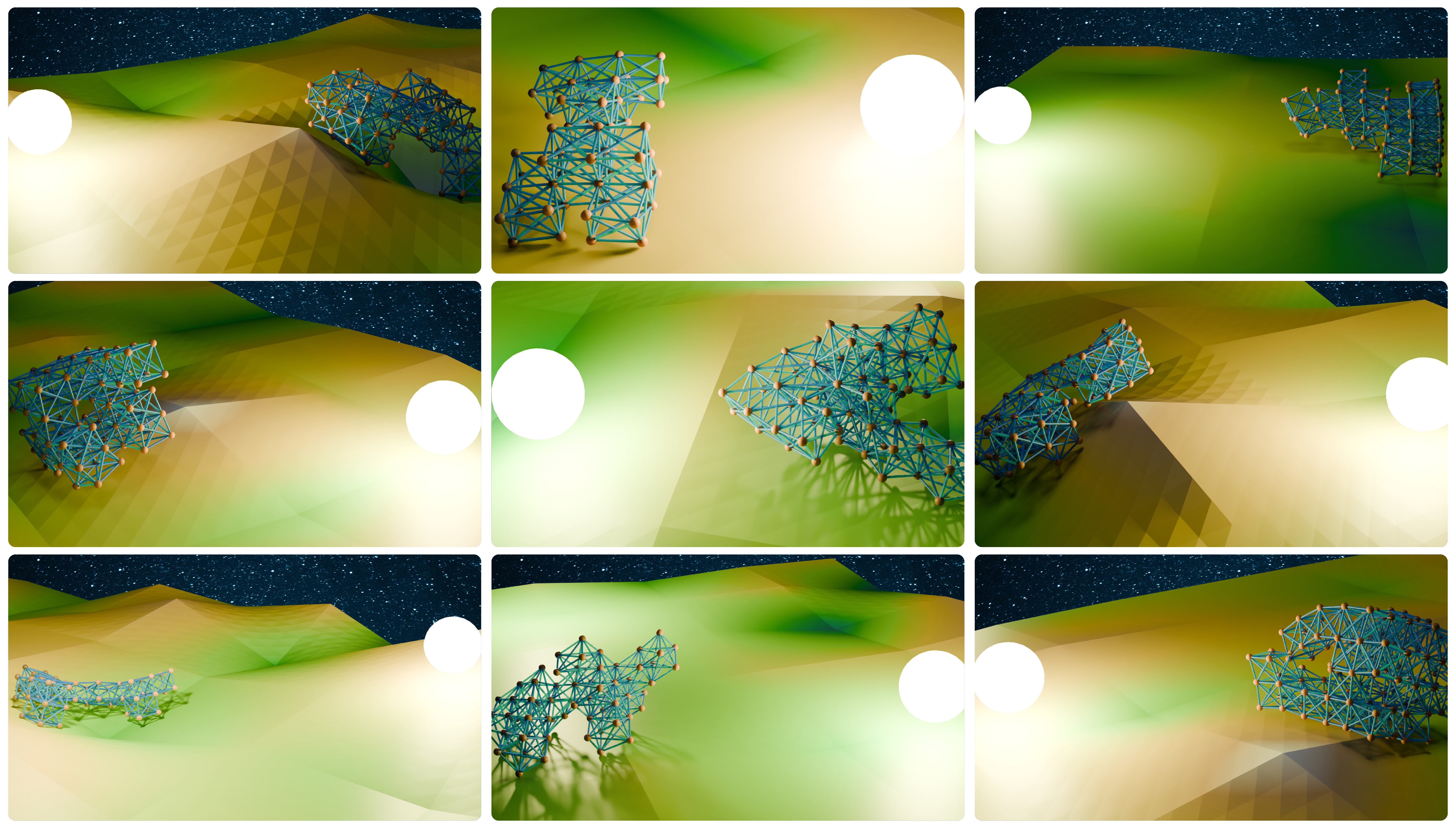}
    \vspace{-20pt}
    \caption{\textbf{Universal control of differentiable robots.}
    Large-scale pretraining and finetuning
    of a universal controller was achieved by
    averaging simulation gradients across the robot's body, world, and goal.
    The controller is shared by an arbitrarily large and morphologically diverse population of robots as
    they undergo morphological evolution.
    The objective is to find designs that can move quickly across a previously-unseen terrain toward a randomly-positioned light source (glowing white spheres).
    }
    \vspace{-10pt}
    \label{fig:intro-robot-teaser}
\end{figure*}

\begin{abstract}

The co-design of robot morphology and neural control typically requires using reinforcement learning to approximate a unique control policy gradient for each body plan, demanding massive amounts of training data to measure the performance of each design. Here we show that a universal, morphology-agnostic controller can be rapidly and directly obtained by gradient-based optimization through differentiable simulation. This process of morphological pretraining allows the designer to explore non-differentiable changes to a robot's physical layout (e.g.~adding, removing and recombining discrete body parts) and immediately determine which revisions are beneficial and which are deleterious using the pretrained model. We term this process ``zero-shot evolution'' and compare it with the simultaneous co-optimization of a universal controller alongside an evolving design population. We find the latter results in \textit{diversity collapse}, a previously unknown pathology whereby the population---and thus the controller's training data---converges to similar designs that are easier to steer with a shared universal controller. We show that zero-shot evolution with a pretrained controller quickly yields a diversity of highly performant designs, and by fine-tuning the pretrained controller on the current population throughout evolution, diversity is not only preserved but significantly increased as superior performance is achieved.

\end{abstract}

\section{Introduction}
\label{sec:intro}

The co-design of morphology and control in robots is
important because robots perform better when their physical layout is optimized for their intended niche---%
like a fish out of water, a good body in one domain can obstruct the acquisition of intelligent behavior in another, if it is unable to evolve.
However, over the past three decades of research, despite exponential increases in computing power, surprisingly little tangible progress has been achieved beyond the very first co-designed robots \cite{sims1994competition}.
This stagnation is due in part to the nested complexity of evolving a robot’s morphology and learning a bespoke controller for every morphological variant.
Because controllers are usually optimized in non-differentiable  simulations using reinforcement learning (RL), large amounts of training data are needed to effectively learn a single controller for a single morphology, a cost that is compounded by repeatedly relearning new controllers as the robot's morphology changes throughout evolution.

As a result, the overwhelming majority of prior work 
has been limited to 
small numbers of
morphological simple robots 
that exhibit simple behaviors 
in simple environments.
Even with simplifying assumptions that significantly speed simulation,
such as constraining the design space to infinitely
rigid ``stick figures'' composed of less than a dozen body parts,
there is usually only enough time to explore a few thousand morphologies 
\cite{zhao2020robogrammar,gupta2021embodied,yuan2022transformact}.
Others have relaxed this constraint by considering more flexible bodies composed of many deformable cells \cite{cheney2018scalable,kriegman2020xenobots,li2025generating}, but due in part to the increased computational burden of simulating soft materials, these robots have had lower motoric complexity (fewer independent motors)
and have been much less intelligent (completely unresponsive to external stimuli) 
compared to their rigid bodied counterparts.
Often the robots are restricted to two dimensional worlds \mbox{\cite{medvet2021biodiversity,wang2023preco,strgar2024evolution}.}

\begin{figure}[!hb]
    \centering
    \includegraphics[width=\columnwidth]{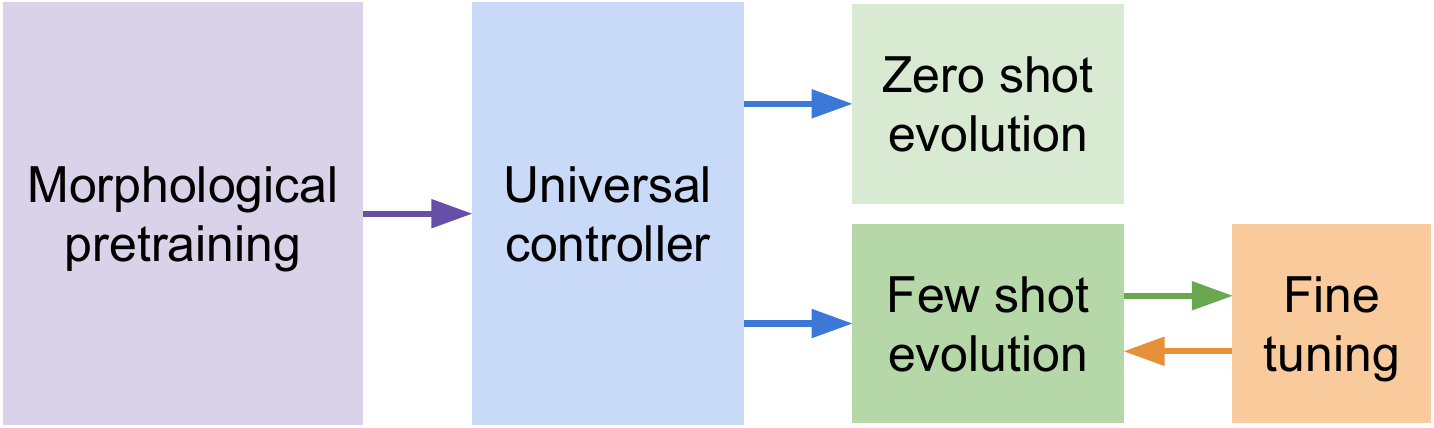}
    \vspace{-16pt}
    \caption{\textbf{Overview of the proposed method.}
    End-to-end differentiable policy training across tens of millions of morphologically distinct robots---morphological pretraining---produces a universal controller, which was kept frozen throughout zero-shot evolution
    and finetuned for each generation of few-shot evolution.
    }
    \vspace{-8pt}
    \label{fig:intro-dataflow}
\end{figure}

\begin{figure*}[t]
    \centering
    \includegraphics[width=\textwidth]{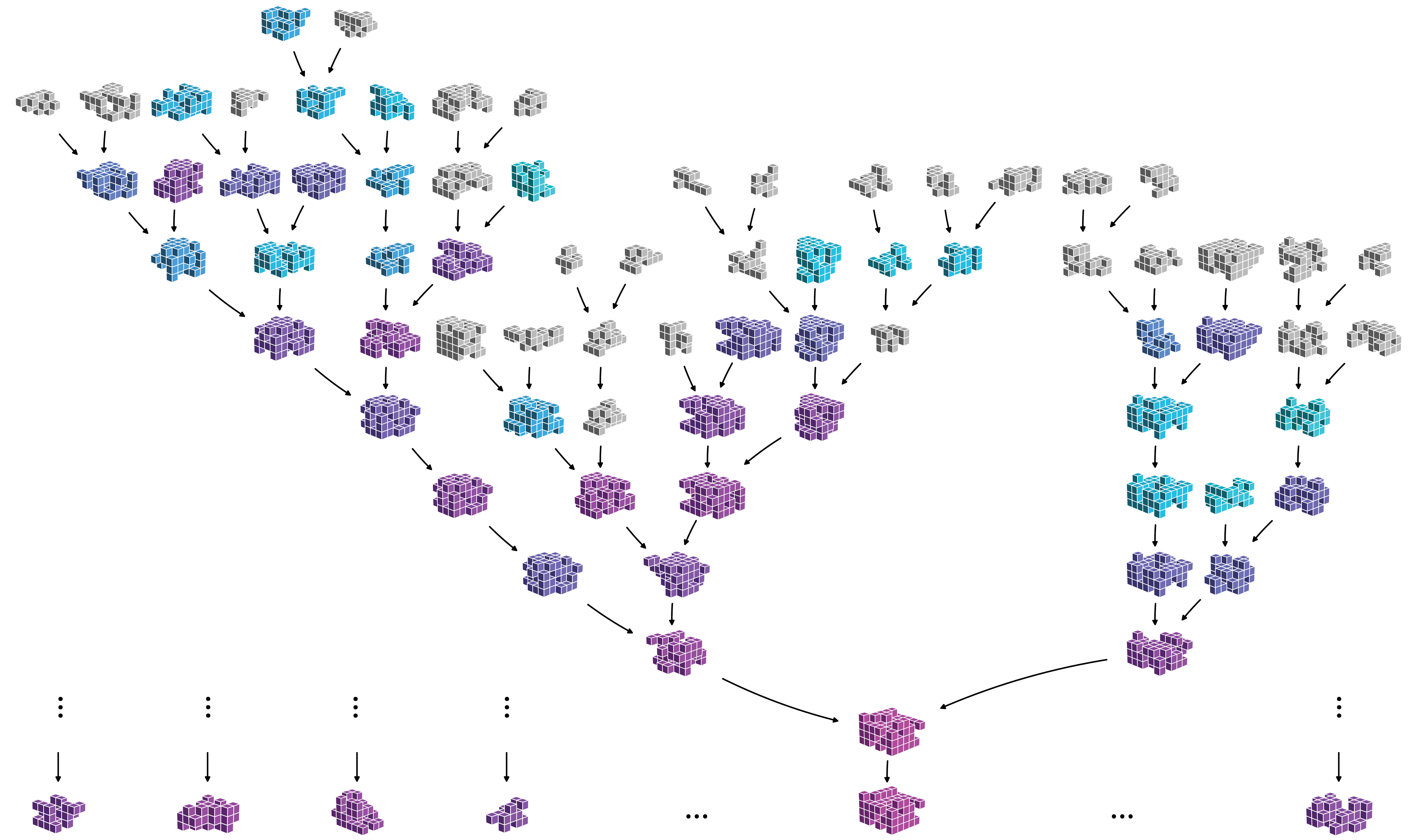}
    \vspace{-14pt}
    \caption{\textbf{Few-shot evolution.}
    A population of 8192 
    initially random designs (a pair of which are shown in the top row)
    were randomly recombined and mutated to produce 8192 offspring, temporarily expanding the population to 16384 designs.
    All designs in the population were driven by the same universal controller, which was rapidly pretrained (before evolution)
    and finetuned for the current population (at every generation of evolution) using analytical gradients
    from differentiable simulation.
    Deleting the worst performing designs and replacing them with the best offspring, and repeating this process for several generations,
    yields a diversity of increasingly performant designs, and ultimately a final population of 8192 winning designs (bottom row), each with their own unique evolutionary history (phylogeny).
    An example phylogenetic tree, colored by loss (decreasing from gray to cyan to pink), is shown for one of winning designs.
    }
    \vspace{-8pt}
    \label{fig:intro-phylogenetic-tree}
\end{figure*}

Inspired by the remarkable success of large-scale pretrained models in computer vision and natural language processing, we here pretrain a universal controller across millions of complex body plans using gradient information from differentiable simulation, 
averaging gradients across variations in the robot's body, world and goal (Fig.~\ref{fig:intro-robot-teaser}).
%
Armed with a universal controller, evolution can now iteratively improve the robot's morphology, 
and the controller can be rapidly finetuned for the current population with simulation gradients (Fig.~\ref{fig:intro-dataflow}).
This also enables the successful recombination of designs (a.k.a.~crossover; Fig.~\ref{fig:intro-phylogenetic-tree}),
a hallmark of biological evolution 
and of human engineering that has yet to be convincingly demonstrated in robots.

Indeed there is a tacit assumption in robotics 
that crossover---%
the combining of two parent designs to produce offspring---%
is so unlikely to produce viable offspring, 
that it is better to omit crossover altogether 
and focus entirely on small mutations 
that slightly alter a single design parent to produce offspring.
While instances have been reported in which two morphologies were combined using crossover to produce a new morphology
\cite{sims1994competition,bongard2001repeated,hiller2010evolving,strgar2024evolution},
it was not clear if crossover ever produced offspring with equal or better fitness than either one of their parents---or if the recombined designs were even better than randomly generated robots.
Here we show how a pretrained universal controller can unlock successful crossover of robots.

Several cases have been reported in the literature in which RL was used to approximate a universal control policy gradient across a small dataset of
previously-designed 
\cite{huang2020one,gupta2022metamorph,bohlinger2024one}
or 
simultaneously co-designed \cite{wang2023preco,li2025generating}
morphologies.
However, the inefficiencies of policy training without recourse to gradient information precluded large scale pretraining.
As we detail below, co-designing morphology and universal control simultaneously from scratch can, and without careful consideration almost certainly will 
result in diversity collapse, inhibiting co-design by reducing it to policy training for a single design.
Others
\cite{ma2021diffaqua,matthews2023efficient,yuhn20234d,cochevelou2023differentiable,strgar2024evolution}
have utilized first-order gradients from differentiable simulation 
to speed co-design.
But, a custom controller still needed to be learned for each morphology,
and the resulting robots could only exhibit rote behaviors,
such as locomotion in a straight line. 

Here we demonstrate a more scalable approach that starts 
with large-scale morphological pretraining 
in differentiable simulation
and yields 
a morphology-agnostic controller 
for adaptive sensor-guided behavior
in complex robots 
with thousands of independent motors.


\begin{figure}[t]
    \centering
    \includegraphics[width=\columnwidth]{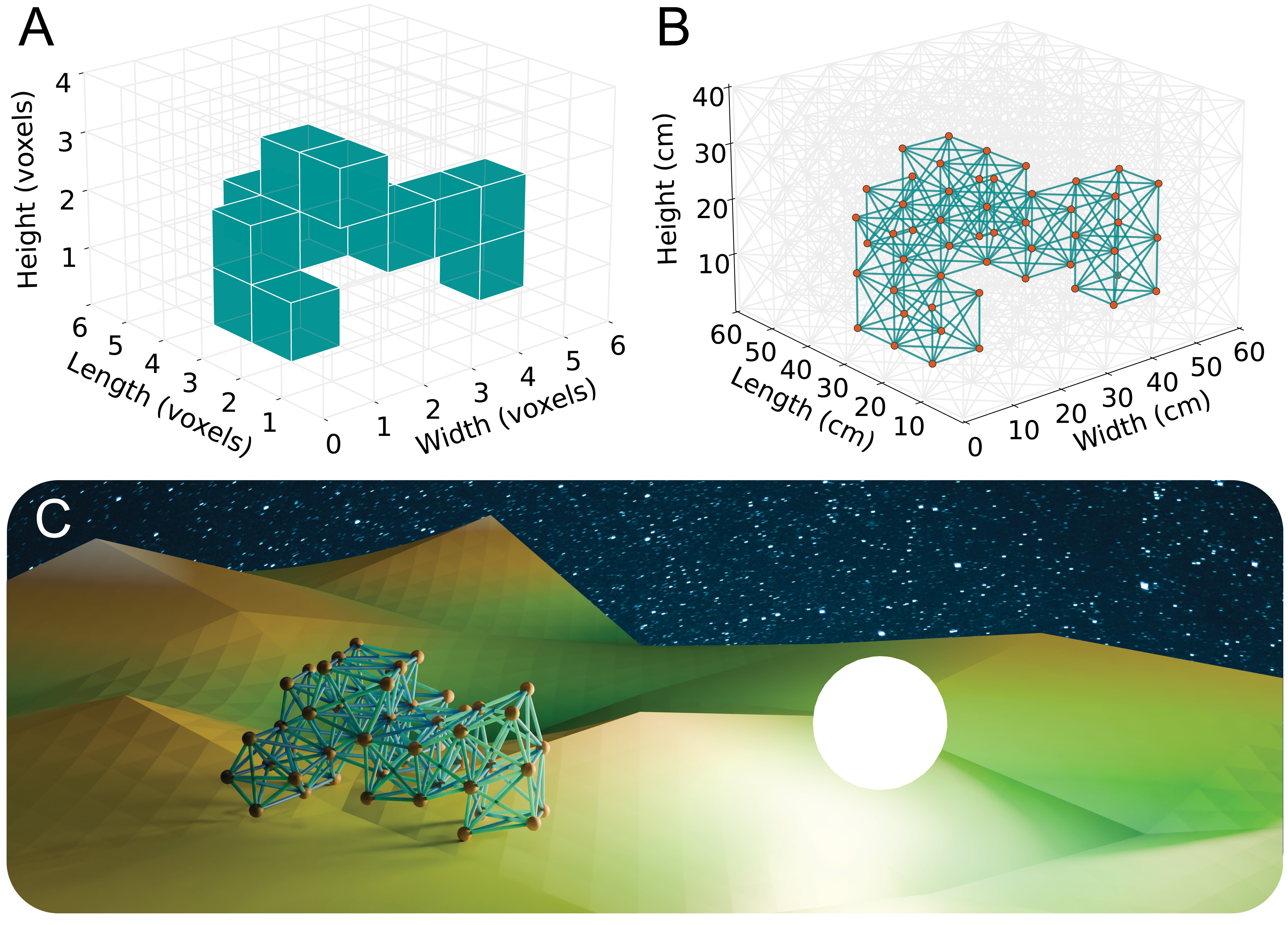}
    \vspace{-18pt}
    \caption{\textbf{Genotype to phenotype.}
    Designs are 
    encoded by voxel genotype (\textbf{A}), 
    which is expressed as a
    spring-mass phenotype (\textbf{B}),
    and evaluated in a
    differentiable environment (\textbf{C}).
    The springs (teal lines in B and C) and masses (small orange spheres) are motorized and sensorized, respectively.
    }
    \vspace{-12pt}
    \label{fig:methods-geno-pheno-sim}
\end{figure}

\section{Methods}
\label{sec:methods} 

In this section we describe 
the morphological design space (the ``morphospace''; Sect.~\ref{methods-morphology-design-space}),
the simulated physical environment (Sect.~\ref{methods-simulator}),
the universal controller (Sect.~\ref{methods-universal-control}),
morphological pretraining (Sect.~\ref{methods-pretraining}),
zero shot evolution (Sect.~\ref{methods-zero-shot-evolution}),
few shot evolution with generational finetuning (Sect.~\ref{methods-few-shot-evolution}),
and 
simultaneous co-design (Sect.~\ref{methods-codesign}).

\subsection{Morphospace}
\label{methods-morphology-design-space}

Robot morphologies were genetically encoded as contiguous collections of voxels 
within a $6 \times 6 \times 4$ (Length $\times$ Width $\times$ Height) binary genotype workspace, $\mathcal{G}$. 
Voxelized genotypes were then mapped to a phenotype space $\mathcal{P}$ comprising 
masses $\mathcal{M}$ and springs $\mathcal{S}$ 
arranged in a cubic lattice 
with 10 cm$^3$ unit cells 
(Fig.~\ref{fig:methods-geno-pheno-sim}).
More specifically, a genotype voxel at position $(i,j,k)$ in $\mathcal{G}$ is expressed phenotypically by eight masses,
one in each corners of the corresponding cubic cell in $\mathcal{P}$ with coordinates $(0.1i+\delta_x,0.1j+\delta_y,0.1k+\delta_z)$ where $\delta_{x,y,z} \in \{0,0.1\}$. 
Springs are then connected to these masses in two patterns: (1) axial springs along cube edges, and (2) planar diagonal springs in each face. 
Adjacent genotype voxels share masses and springs at their interfaces, 
ensuring that contiguous structures in $\mathcal{G}$ mapped to cohesive mass-spring networks in $\mathcal{P}$.

The resultant $6 \times 6 \times 4$ workspace accommodated a maximum of $|\mathcal{M}|=245$ potential mass positions and $|\mathcal{S}|=1648$ potential springs. 
Each robot was centered in the x-y plane according to its center of mass and shifted to the bottom of the workspace to ensure ground contact prior to behavior. 
This procedure ensured stable initial conditions for locomotion while maintaining consistent relative positioning between robots of different morphologies.

To identify unique morphologies, we defined an equivalence relation on the genotype space that accounted for translations and symmetries. Two genotypes were considered identical if, after aligning their occupied voxels to the origin, one could be transformed into the other through any combination of: (1) 90° rotations about the z-axis, (2) reflection about the x-axis, or (3) reflection about the y-axis. Each unique design was represented by its lexicographically minimal form across all such transformations.

\subsection{Differentiable simulation}
\label{methods-simulator}

We here extend the differentiable 2D mass-springs simulators developed by \citet{hu2019difftaichi} and \cite{strgar2024evolution} to three dimensions and add exterception: perception of external stimuli outside the body, namely light.
Masses on $\mathcal{M}$ hosting photoreceptors were connected by actuating springs on $\mathcal{S}$ (defined above in Sect.~\ref{methods-morphology-design-space}), which exerted forces on their endpoint masses to perform phototaxis: movement toward a light source. 

During simulation, 
spring rest lengths may be actuated continuously between $\pm 20\%$ of their initial values derived from $\mathcal{P}$ (see Sect.~\ref{methods-morphology-design-space}). 
Spring forces were computed according to Hooke's law $F = k(L - L_0)$, where $k=1.5 \times 10^4$ N/m is the spring stiffness coefficient, $L$ is the current spring length, and $L_0$ is the modulated rest length. 
Resulting impulses, as well as damping and gravitational forces, were used to update velocities for each mass, and in turn mass positions were updated using the new velocities.

The terrains along which robots behaved were modeled using randomly sampled height maps (see Appx.~\ref{appendix-dataset-random-environment-generation} for details). 
During simulation, terrain heights at arbitrary coordinates $(x, y)$ were computed through bilinear interpolation of the height map. 
For collision handling, we detected when a mass' updated $z$-coordinate fell below the interpolated terrain height at its $(x, y)$ position. Upon detection, we employed a three-phase resolution: (1) iterative bisection on the interval [0, $dt$] to estimate the time of impact and advance the mass to the contact point, (2) velocity projection onto the contact surface normal (estimated via central differences), and (3) constrained motion along the surface tangent for the remaining timestep. Following \citet{strgar2024evolution}, friction forces were computed by negating the tangential velocity component and clamping its magnitude to not exceed the magnitude of the normal velocity component.

Our simulator was implemented in the Taichi programming language \cite{hu2019difftaichi}, providing both GPU acceleration for parallel, multi-robot simulation and automatic differentiation capabilities. The simulator was directly integrated with a PyTorch-based universal controller (Sect.~\ref{methods-universal-control}), enabling end-to-end backpropagation through 1000 timesteps ($dt = 0.004$s) of physics simulation and neural control for gradient-based optimization of the controller parameters.

\begin{figure}[b]
    \centering
    \vspace{-10pt}
    \includegraphics[width=\columnwidth]{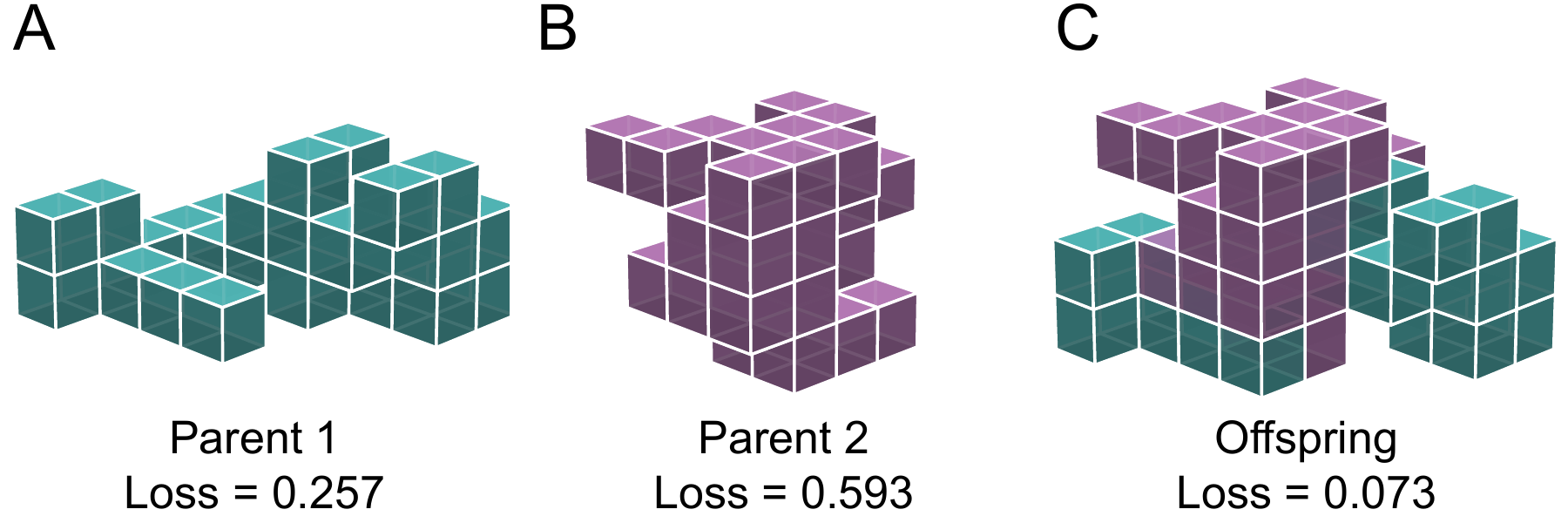}
    \vspace{-18pt}
    \caption{\textbf{Recombination of substructures.}
    A pair of designs (parents; \textbf{A, B}) is combined via crossover to produce a new design  (offspring; \textbf{C}) that inherits components from both parents.
}
    \label{fig:results-xover}
    \vspace{-8pt}
\end{figure}
\begin{figure*}[t]
    \centering
    \includegraphics[width=0.975\textwidth]{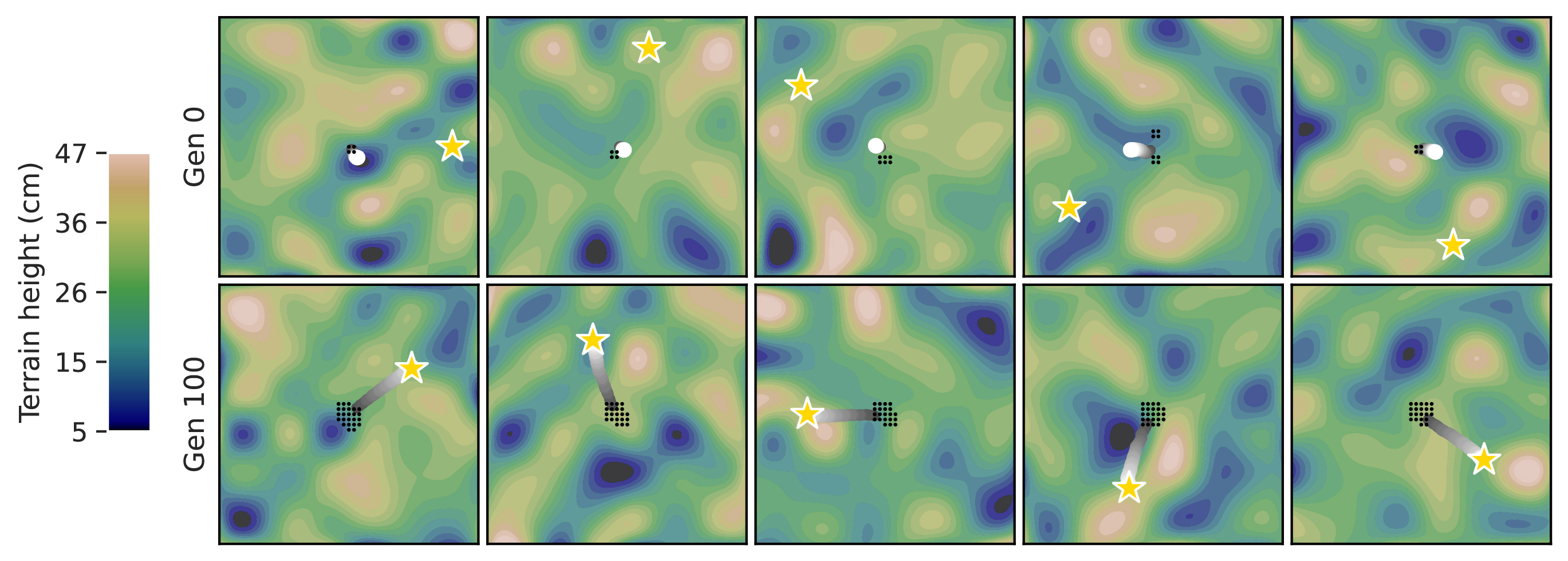}
    \vspace{-10pt}
    \caption{\textbf{Evolution of phototaxis.}
    The five worst designs in the population are depicted before (top row) and after (bottom row) zero-shot evolution.
    Each design (black dotted footprints) was placed in the center of a randomly generated map.
    Before evolution,
    not all of designs in the population could
    move (gray to white trajectories)
    across any terrain
    toward a light source (gold stars) using the pretrained controller.
    After evolution, they could.
    One of the design principles that evolution discovered is that larger footprints increase locomotion stability.
    }
    \vspace{-12pt}
    \label{fig:results-com-traces}
\end{figure*}
\begin{figure*}[t]
    \centering
    \includegraphics[width=0.95\textwidth]{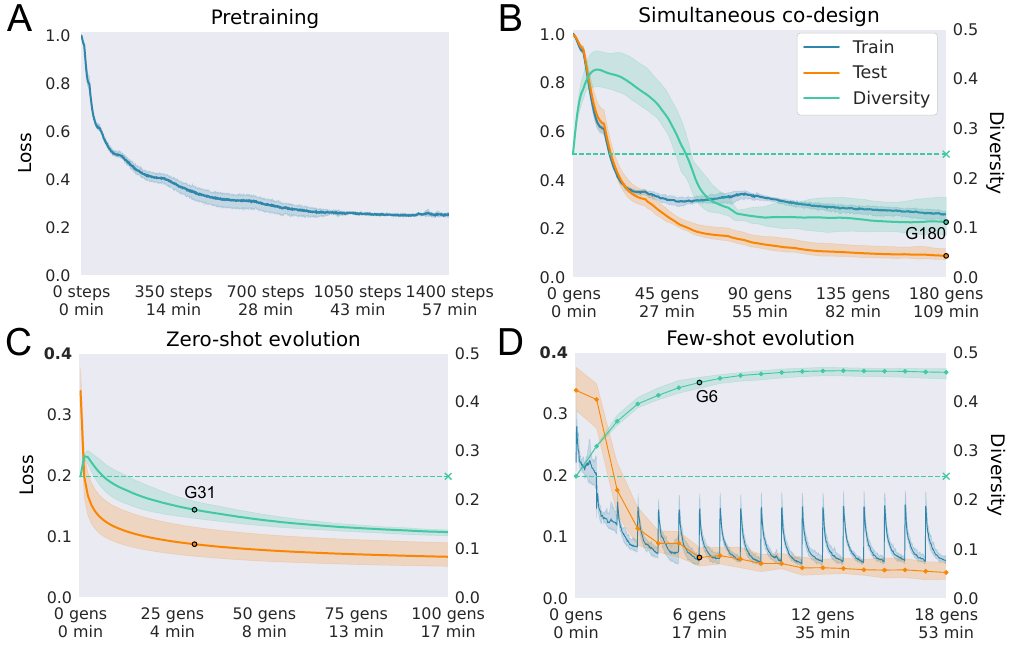}
    \vspace{-8pt}
    \caption{\textbf{Performance and diversity.} 
    Morphological pretraining (\textbf{A}) converges with 70\% improvement from baseline. 
    Simultaneous co-design (from scratch without pretraining; \textbf{B}) achieves similar training loss; 
    but, population diversity (mean pairwise Hamming distance on genotypes) collapses as evolution converges to a single species of similar designs which simplifies shared control.
    Zero shot evolution (using the pretrained controller; \textbf{C}) rapidly improves test performance, but also suffers diversity collapse as evolution compiles slightly modified clones of the designs that are the most compatible with the pretrained model.
    Few-shot evolution (\textbf{D}) resets the pretrained controller at the start of each generation 
    and performs 60 finetuning steps per generation. 
    This significantly increases and sustains diversity as well as performance. 
    Solid lines represent 
    the batch (training; blue)
    or
    population (test; orange) means, 
    averaged across three independent trials.
    Shaded regions surrounding the mean curves show the minimum and maximum values across the three trials.
    }
    \vspace{-10pt}
    \label{fig:results-performance}
\end{figure*}

\subsection{The universal controller}
\label{methods-universal-control}

We employed a simple multi-layer perceptron (MLP) as a universal controller for adaptive phototaxis: guiding a population of thousands of morphologically diverse robots towards arbitrarily positioned light sources across randomly varying, rugged terrains.
The network mapped two input streams to spring actuation signals: photosensor readings from masses and central pattern generator (CPG) inputs. 
To accommodate all possible body plans in the morphospace (defined in Sect.~\ref{methods-morphology-design-space}), the network's input dimension was set to $|\mathcal{M}|$ (the maximum number of masses) and output dimension to $|\mathcal{S}|$ (the maximum number of springs). Sensors and actuators not present in the a specific robot's body had their corresponding signals masked to zero, providing an implicit morphological conditioning through observation and action space masking.

Each mass-bound photosensor measured light intensity following the inverse square law relative to the light source position. 
Sensor readings for each robot were normalized by subtracting the mean computed across that robot's active (unmasked) sensors, providing a zero-centered, embodied irradiance gradient. 

Following \citet{hu2019difftaichi} and \citet{strgar2024evolution},
CPG inputs consisted of five sinusoidal waves with angular frequency $\omega=10$ rad/s and phase offsets evenly spaced by $2\pi/5$ radians. Over the 4-second simulation period (1000 timesteps, $\Delta t=4e-3$s), these oscillators completed approximately six cycles.

The MLP architecture consisted of an input layer (dimension 250: $|\mathcal{M}|$ mass sensors plus 5 CPG inputs), three hidden layers (dimension 256 each), and an output layer (dimension 1648: $|\mathcal{S}|$ springs). Each hidden layer was followed by layer normalization and ReLU activation, while the output layer used a $\tanh$ activation. All layers included learnable biases. In total the model consisted of 620,912 learnable parameters. 

Network weights were initialized using a Xavier uniform distribution (gain=1.0) \cite{glorot2010understanding} with zero-initialized biases, and the network was optimized using Adam \cite{adamkingma} ($\beta_1=0.9$, $\beta_2=0.999$) with gradient norm clipping at 1.0. 
Learning rates were scheduled using variants of cosine annealing with restarts (detailed in Sects.~\ref{methods-pretraining},~\ref{methods-few-shot-evolution}, and~\ref{methods-codesign}).

\subsection{Morphological pretraining}
\label{methods-pretraining}

The universal controller was pretrained 
across a dataset of over 10 million distinct robot morphologies 
(see Appx.~\ref{appendix-dataset-random-robot-generation} for details). 
The controller was trained over 1400 learning steps to minimize the batch mean of $d_1/d_0$, where $d_1$ and $d_0$ represent each robot's final and initial distances from its target light source, respectively. 
This relative distance formulation ensured robots were not penalized for being initialized far from their targets and equally incentivized fine-grained control in robots initialized near their targets.

We used a batch size of 8192, distributed in equal partitions of 1024 across a single compute node consisting of eight H100 SXM GPUs. 
Each sample consisted of a randomly-generated robot morphology (see Appx.~\ref{appendix-dataset-random-robot-generation} for details) 
a randomly-generated terrain shape
and a randomly-positioned light source (see Appx.~\ref{appendix-dataset-random-environment-generation} for details), and was seen exactly once during training. 
Training used a cosine annealing with restarts schedule, with initial learning rate $1e^{-3}$, cycle length starting at 10 steps and doubling each restart, minimum learning rate $1e^{-5}$, and a decay rate of 0.7 applied to the starting learning rate at each cycle. 

\subsection{Zero-shot evolution}
\label{methods-zero-shot-evolution}

Here, we introduce a novel robot design paradigm that leverages a frozen, pretrained universal controller to rapidly evaluate non-differentiable changes to a given robot's body plan. 
By using a single, fixed controller for all body plans,
the design space may be efficiently explored without the computational burden of training a custom controller for each body plan. 
We refer to this method as ``zero-shot evolution''. 

We initialized a population of 8192 random robot morphologies (unseen during pretraining) and evaluated each on a fixed test set of terrain and light source position pairs (see Appx.~\ref{appendix-dataset-evaluation-environments} for details). A simple genetic algorithm was then applied iteratively: the population produced an equal number of offspring through two variation operators (described below), new offspring were evaluated once on the test set, and the top 50\% across parents and offspring (using cached evaluation scores for parents) were selected to form the next generation.

Robot offspring were produced through one of two variation operators: mutation and recombination. The population was partitioned into two distinct groups: a random 25\% of members were assigned to produce offspring through mutation, while the remaining 75\% were reserved for producing offspring through recombination (or crossover). Each member in the mutation group produced a single offspring through random bit flip mutations performed on their genotype. Flips occurred with probability $p = 1/N$ where $N = 6 \times 6 \times 4$, the total number of voxels in the robot's genotype. After mutation, genotypes were processed to ensure validity: only the largest connected component was retained, and the resulting structure was translated to the bottom center of the workspace. If a mutation produced a body that was either empty or identical to a previously seen body, the process was repeated with the mutation rate increased by 2.5\% until a valid, unique design was obtained.

From the recombination group (75\% of the population), pairs of distinct parents were randomly sampled to produce offspring through crossover (Fig.~\ref{fig:results-xover}). 
For each sampled pair, an offspring's genotype was created using a bitwise exclusive or (XOR) operation on the parent genotypes. As with mutation, post-processing retained only the largest connected component and centered it at the bottom of the workspace. If the resulting design duplicated a previous one, it was discarded. The sampling and generation process was repeated until the number of offspring equaled the size of the recombination group (75\% of the population).

\subsection{Few-shot evolution}
\label{methods-few-shot-evolution}

In this experiment we extend the zero-shot paradigm (described above in Sect.~\ref{methods-zero-shot-evolution}) by fine-tuning 
the pretrained universal controller 
to the current population
at every generation of morphological evolution.
We refer to this approach as ``few-shot evolution''.
The experimental setup of few-shot evolution matched the zero-shot case, with one key difference: before evaluation, each generation received 60 fine-tuning steps (30 for parents, 30 for offspring).
The number of fine-tuning steps was
empirically chosen to balance controller adaptation against evolutionary search while maintaining comparable maximum wall-clock time across experiments. 
At the start of each generation, the controller's weights were reset to their pretrained values and the optimizer state was reinitialized. 
Fine-tuning used a cosine annealing learning rate schedule with initial and minimum rates of $3.5e^{-4}$ and $3.5e^{-5}$, respectively. The cycle length was set to 100; however each cycle was truncated to align each cycle with one generation's 60 fine-tuning steps resulting in an effective minimum learning rate of $1.5e^{-4}$. 
Since every generation re-initialized the pretrained weights, we did not decay the learning rate at the start of each cycle. 

\subsection{Simultaneous co-design from scratch}
\label{methods-codesign}

In our third and final experimental group, 
we remove morphological pretraining
and instead 
simultaneously 
evolve a population of robots
and
learn their universal controller, 
from scratch.
Unlike few-shot evolution, controller parameters and optimizer state are inherited across generations rather than being reset. The genetic algorithm operates as before, but we reduce the per-generation training to just 2 learning steps (1 for parents, 1 for offspring) to maintain parity with our pretraining experiments, where each training batch was unique.

Initially, we employed the same cosine annealing learning rate schedule used in morphological pretraining, but we found it was beneficial to reduce the start-of-cycle learning rate decay factor from 0.7 to 0.65 in order to stabilize learning across cycle restarts in this setting.

\section{Results}
\label{sec:results}

In this section we evaluate 
the results of 
morphological pretraining (Sect.~\ref{results-pretraining-performance}),
zero- and few shot evolution (using the pretrained model; Sect.~\ref{results-zero-and-few-shot-evolution}),
and simultaneous co-design from scratch (without pretraining; Sect.~\ref{results-simultaneous-co-design}).

\subsection{Pretraining performance}
\label{results-pretraining-performance}

Across three independent trials,
each using a distinct dataset of 
randomly-generated morphologies and environments,
pretraining 
exhibited stable learning trajectories 
with low variance across trials (Fig.~\ref{fig:results-performance}A), 
converging in approximately 1,400 learning steps (56 minutes of wall-clock time).
Loss was defined as the ratio of final to initial distance from the target light source. 
At initialization with random controller weights, this ratio was 1.0, indicating robots remained stationary throughout simulation. 
After pretraining, the loss stabilized at approximately 0.3, representing a 70\% improvement. 
That is, in environments sampled from the training distribution, robots using the pretrained universal controller traversed an average of 70\% of their initial distance to the light source. 
Since each training batch used novel morphologies, we omitted model selection with a \mbox{validation set.}


To visualize the breadth of morphological diversity handled by the pretrained controller, Fig.~\ref{fig:appendix-zero-shot-robot-grid} showcases a representative sample of successful robots. 
These examples were selected uniformly from the top-performing 50\% of the test morphology set. 
The selected bodies exhibit high variation in both scale and morphological characteristics demonstrating the non-trivial generalization of the universal controller.

\begin{figure}[t]
    \centering
    \includegraphics[width=\columnwidth]{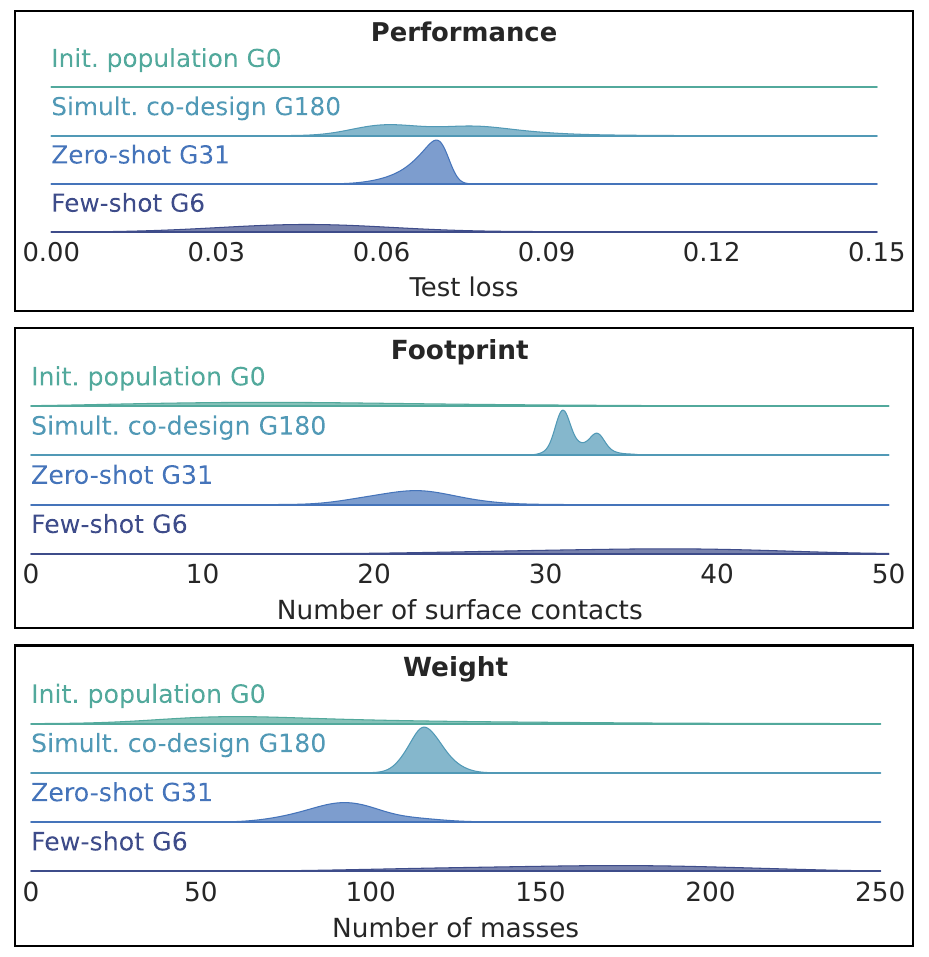}
    \vspace{-20pt}
    \caption{\textbf{Evolved populations.}
    Population performance, phenotype footprint size, and body mass for the initial (randomly generated) and evolved design populations.
    }
    \vspace{-18pt}
    \label{fig:appendix-robot-data-gen-stats}
\end{figure}

\subsection{Zero- and few shot evolution (with pretraining)}
\label{results-zero-and-few-shot-evolution}

A population of morphologies was evolved through random mutation and crossover operations, using the pretrained universal controller.
On the same challenging set of tasks used for evaluating pretrained controller generalization,
the population converges to near optimal performance
in 100 generations of evolution (17 minutes of wall-clock time)
without finetuning the controller (``zero shot evolution''; Sect.~\ref{methods-zero-shot-evolution}).
Although zero-shot evolution shows rapid convergence in controlling thousands of distinct bodies, this success masks a key pattern: design population diversity decreases as performance improves. 
Fig.~\ref{fig:results-performance}C reveals this pattern---after a brief diversity spike at evolution's onset, the population gradually homogenizes. 
We term this phenomenon diversity collapse, measuring diversity as the population's mean, pairwise Hamming distance in (and normalized to) the genotype space $\mathcal{G}$ (defined in Sect.~\ref{methods-morphology-design-space}). This metric naturally reflects differences in morphology (body) as well as sensing and actuation masking in the universal controller (brain).

We found that generational finetuning of the universal controller for the current population (``few shot evolution'')
not only preserves diversity but in fact significantly increases diversity (Fig.~\ref{fig:results-performance}D).
This is a somewhat surprising result as there was no explicit selection pressure to maintain diversity.
The process of morphological evolution seems to intrinsically increase population diversity. 
However, in absence of generational finetuning, there is a tipping point at which it is easier to purge diversity, replacing the worst designs with slightly modified clones of the best, than to discover novel morphological innovations with superior performance.

\subsection{Simultaneous co-design (\textit{without} pretraining)}
\label{results-simultaneous-co-design}

Ablating pretraining (and funetuning),
and instead simultaneously optimizing morphology and universal control, together from scratch, 
results once again in rapid diversity collapse (Fig.~\ref{fig:results-performance}B).
Performance plateaus in well under 180 generations, corresponding to 360 controller learning steps and 109 minutes of wall-clock time.
The extent of diversity collapse can be seen in Fig.~\ref{fig:results-morpho-grid}B, where we visualize morphologies from one of the three independent trials,
and in Fig.~\ref{fig:appendix-robot-data-gen-stats} where we plot morphological variance across evolved populations in terms of footprint size and body weight.

In all three co-design paradigms (zero shot, few shot, simultaneous), universal control enabled successful crossover (Fig.~\ref{fig:appendix-variation-operator-stats}).
In terms of offspring survival,
crossover was initially much more successful than mutation.
But in the case of simultaneous co-design, this was not an apples to apples comparison because each generation provided the controller with more time to learn how to control the population, and the randomly initialized controller was very bad at the task.
And so it was not clear if the success of offspring was due to changes in parent morphology or improvements to the universal controller.
The superior performance of pretraining across random morphologies, shows that the designs produced by crossover during simultaneous co-design were no better than random designs. 
In zero-shot and few-shot evolution, however, the pretrained controller is quite good at the very start, and in zero-shot the controller is not updated during evolution, providing clear evidence of successful crossover prior to diversity collapse.

\section{Discussion}
\label{sec:discussion}

In this paper, we introduced 
the large-scale pretraining and finetuning of a universal controller using differentiable simulation
and demonstrated how this approach
accelerates the 
design
of complex robots.
The learned controller 
allows most 
randomly-generated morphologies (mass-spring networks)
to orient along a randomly-generated stimulus (light) vector in three dimensions, 
and to follow the vector to its source (phototaxis)
across challenging, randomly-generated environments (terrains)---more or less: 
some designs were much better
than others, and some outright fail (Fig.~\ref{fig:results-com-traces}).
Using the pretrained model as a prior, the designer 
can quickly
explore a diversity of changes---from subtle mutations to large recombinations---across
arbitrary numbers of
distinct designs in parallel
without destroying the functionality of working designs, and without constantly readapting the controller to support every morphological innovation.

We intentionally chose a vanilla evolutionary algorithm 
as ``the designer'' 
and a minimal neural architecture for the universal controller
to illustrate the power and potential of our approach.
We were particularly surprised by the effectiveness of a simple MLP in controlling 
such large numbers of morphologically complex robots 
across such challenging terrains.
Interestingly, the gaits generated by the universal controller were quite different from those 
tailored for individual body plans
in similar conditions 
\cite{strgar2024evolution};
instead of walking or ambling across the rugged terrain,
the universal controller
discovered patterns of saltation
(hopping) not unlike that of kangaroos, in which coordinated actuation of muscles is followed by an aerial phase.

It is important to note, however, that while this controller was universal across the robot's morphology and task environment, 
we only considered 
a single
material (soft),
percept (light),
actuator (linear),
and task (phototaxis).
Extending this approach to multiple tasks 
that demand 
more intricate, multi-material body plans with
multi-modal sensing
(e.g.~not just moving toward a single stimulus source, 
but reacting to various other stimuli, 
manipulating objects, 
and working with or against other robots...)
may require gradually complexifying the neural architecture.
This will likely also require replacing the direct genotype-to-phenotype mapping with
more a sophisticated (pleiotropic)
compression of phenotypes 
into a latent genome
\cite{li2025generating}.
Instead of presupposing voxel cells with
two dozen springs and eight masses,
latent genes could control the expression of more atomic building blocks,
such as individual masses and springs (or subatomic particles within them),
allowing other kinds of non-cubic cells \cite{hummer2024noncubic} to emerge.
If extended to self-reconfigurable robots, the latent genome or many such genomes may be expressed in myriad ways by a single robot with universal self control.

We also identified in this paper a previously unknown yet inherent problem of co-designing morphology and universal control---diversity collapse---%
and showed how to solve this problem through
generational finetuning.
However, this first investigation of diversity collapse 
only considered a single measure of morphological diversity.
Other metrics at both the morphological and behavioral level could be formulated or derived from a latent genotype space.
Such metrics could then be incorporated into the design algorithm as a constraint or additional objective.

Another important limitation of this work was that the simulated designs were not transferred to reality.
Doing so may require
higher resolution simulations (Fig.~\ref{fig:results-dog})
or
improvements to the simulator, e.g. it's model of contact, light, and light sensors.
Adding noise to these models can also ensure that the robot's behavior does not exploit inaccuracies of the simulation \cite{jakobi1995noise}.
Or the simulator could be augmented with a neural network that learns the residual physics that were not accounted for a priori \cite{gao2024sim}.
However, the universal controller itself might help reduce the simulation-reality gap since it is already by definition insensitive to a wide range of variation in the simulated robot's body and world.

Despite these limitations, the sheer scale and efficiency achieved by this work
opens a new frontier in robot co-design through automatic differentiation, 
suggesting the breadth of infrastructure and theory developed in fields of deep learning and neural networks may be leveraged by robot co-design in future work.

\section*{Acknowledgments}
This research was supported by
NSF award FRR-2331581
and
Schmidt Sciences AI2050 grant G-22-64506.

\bibliography{main}
\bibliographystyle{icml2025}

\clearpage
\appendix
\onecolumn
\section{Appendix}


\subsection{Random robot generation}
\label{appendix-dataset-random-robot-generation}

\begin{figure*}[t]
    \centering
    \includegraphics[width=\textwidth,keepaspectratio=true]{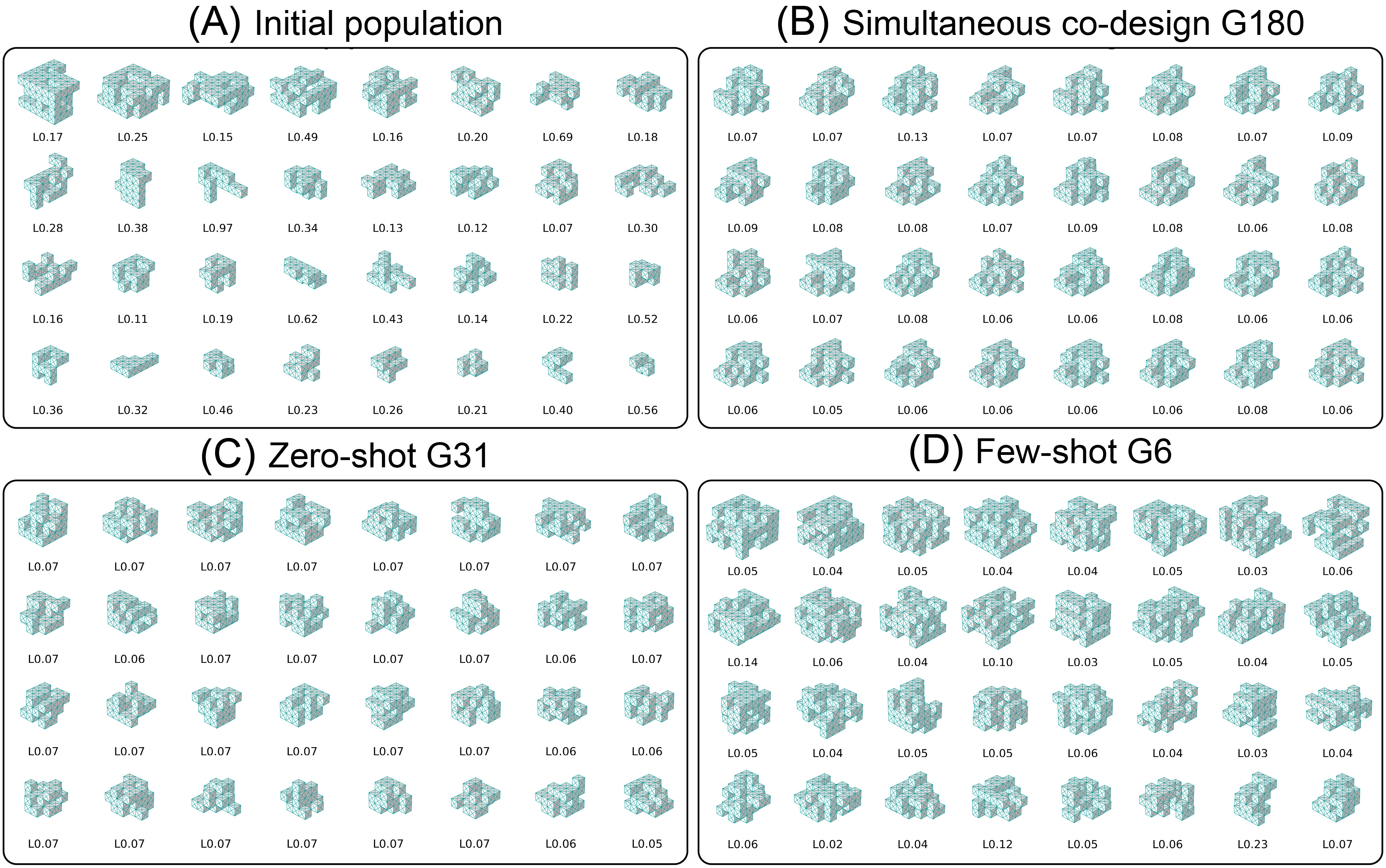}
    \vspace{-16pt}
    \caption{\textbf{Morphological distinctiveness.}
    Robot designs shown are sampled uniformly from each generation's test performance distribution and arranged (left to right, top to bottom) by morphological distinctiveness, defined as the mean pairwise Hamming distance to its peer designs. Performance scores appear below each design. The initial population (\textbf{A}) exhibits diverse morphologies with broad performance variation, serving as the starting point for all methods. After 180 generations, simultaneous co-design (\textbf{B}) yields high-performing but morphologically homogeneous designs. In contrast, both zero-shot evolution at generation 31 (\textbf{C}) and few-shot evolution at generation 6 (\textbf{D}) achieve equal or superior performance while maintaining greater morphological diversity and complexity.
    } 
    \vspace{-8pt}
    \label{fig:results-morpho-grid}
\end{figure*}

\begin{figure*}[t]
    \centering
    \includegraphics[width=\textwidth]{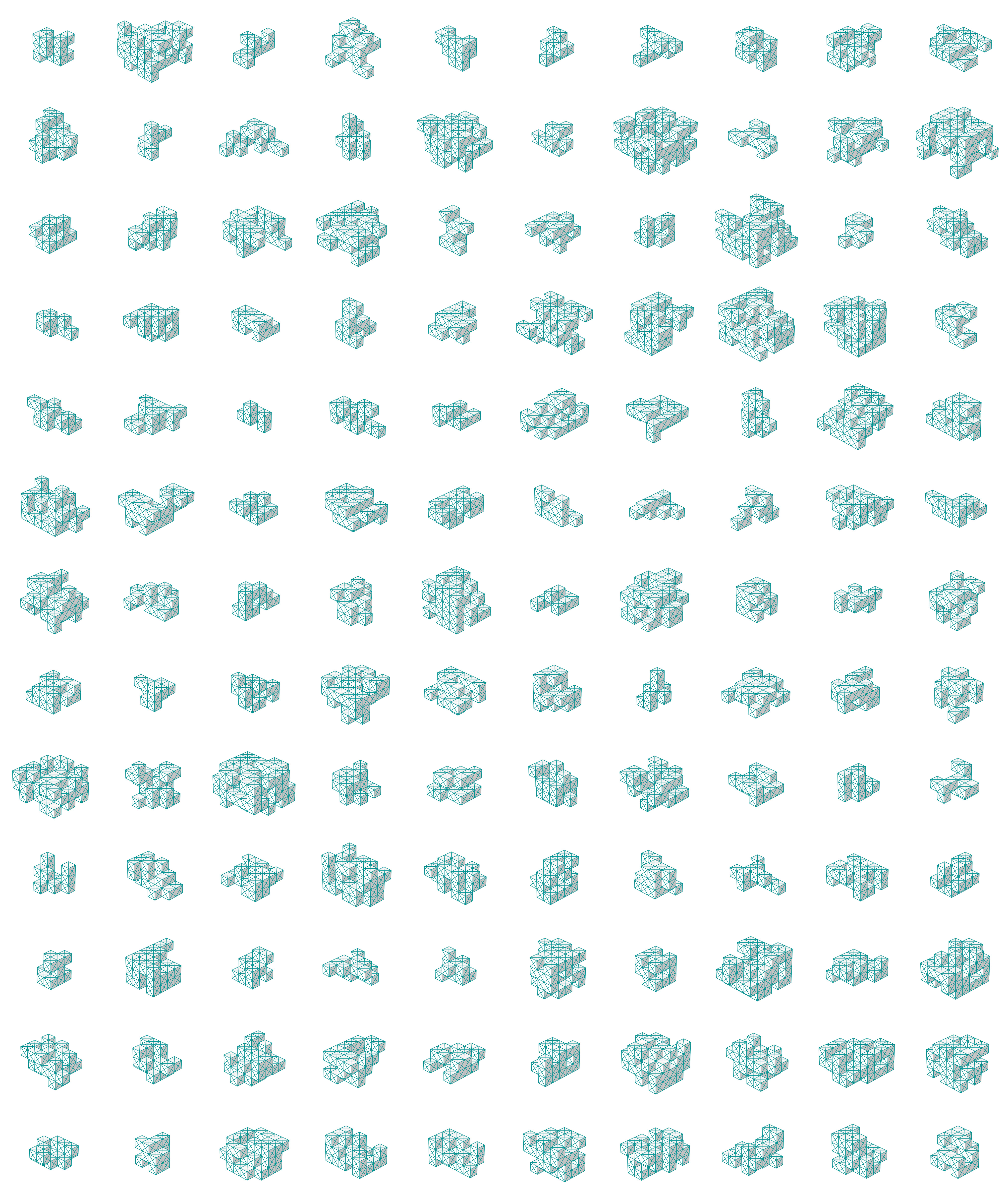}
    \vspace{-18pt}
    \caption{\textbf{Generalization of pretrained universal controller.} Randomly sampled morphologies from the top 50\% of performers in generation 0 of zero-shot evolution. The universal controller successfully controls these diverse, previously unseen robot designs, demonstrating effective generalization across morphologies.}
    \vspace{-8pt}
    \label{fig:appendix-zero-shot-robot-grid}
\end{figure*}

Random morphologies were generated de novo during pretraining and as the initial seed population of evolution (gen 0). 
First, we enumerated all possible (length, width, height) tuples with length and width in $[1, 6]$ and height in $[1, 4]$, corresponding to the voxel dimensions of $\mathcal{G}$ (see Sect.~\ref{methods-morphology-design-space}). 
We then randomly sampled a volume uniformly from the set of possible volumes and subsequently sampled a compatible (length, width, height) tuple to define the bounding box for the genotype. 
This ensured our dataset contained morphologies of varying volumes and dimensions. 
Within this bounding box, all voxels were initialized as inactive (zero) and then randomly activated according to probability $p \sim \mathcal{N}(\mu=0.35, \sigma=0.125)$, clipped to $[0.1, 0.6]$.  
If the resulting structure contained no active voxels, sampling was repeated. The largest connected component of active voxels was retained to ensure a valid morphology. 
If necessary, the bounding box was zero-padded back to $6 \times 6 \times 4$ and the connected component was centered in the horizontal plane and shifted to the bottom of the workspace. 
The parameters of the sampling distribution were empirically set to produce diverse structures, a sampling of which can be visualized in Figs.~\ref{fig:results-morpho-grid}A
and 
\ref{fig:appendix-zero-shot-robot-grid}.

\subsection{Random environment generation}
\label{appendix-dataset-random-environment-generation}

Random environments were generated during pretraining (Sect.~\ref{methods-pretraining}), 
few-shot evolution (Sect.~\ref{methods-few-shot-evolution}) 
and 
simultaneous co-design (Sect.~\ref{methods-codesign}). Zero-shot evolution did not require random environment generation since there was no model training involved and thus relied only on evaluation environments (Appx.~\ref{appendix-dataset-evaluation-environments}).
An environment consisted of a (terrain, light source position) tuple. A random terrain was generated by sampling a discrete $8 \times 8$ height map of uniformly spaced values. Each value was sampled independently from a Gaussian distribution $\mathcal{N}(\mu = 0, \sigma = \mathcal{U}(0, 0.1))$. A light source position was generated by sampling $(x, y)$ coordinates uniformly inside the circle $(x - r_x)^2 + (y - r_y)^2 = r^2$, where $r \sim \mathcal{U}(0.4, 2.0)$ and ($r_x$, $r_y$) was the initial center of mass position of each robot. Prior to the start of simulation light source positions were placed in 3D by incorporating the terrain height at the sampled (x, y) location.

\subsection{Evaluation environments}
\label{appendix-dataset-evaluation-environments}

Each generation of zero-shot evolution (Sect.~\ref{methods-zero-shot-evolution}), 
few-shot evolution (Sect.~\ref{methods-few-shot-evolution}) 
and simultaneous co-design (Sect.~\ref{methods-codesign}),
both the parents and their offspring were evaluated on a fixed set of 10 testing environments. 
This dataset was constructed as follows. 
Light sources were placed in two rings centered about the robot's starting position: five targets at radius 1.5 and five at radius 2.0, with their angular positions offset to maximize radial coverage. 
Terrains were sampled at five uniformly spaced difficulty levels, characterized by height map standard deviations in $\{0.02, 0.04, 0.06, 0.08, 0.1\}$. 
Each ring of light position targets was randomly paired with one terrain from each difficulty level.

\begin{figure*}[b]
    \centering
    \includegraphics[width=\textwidth]{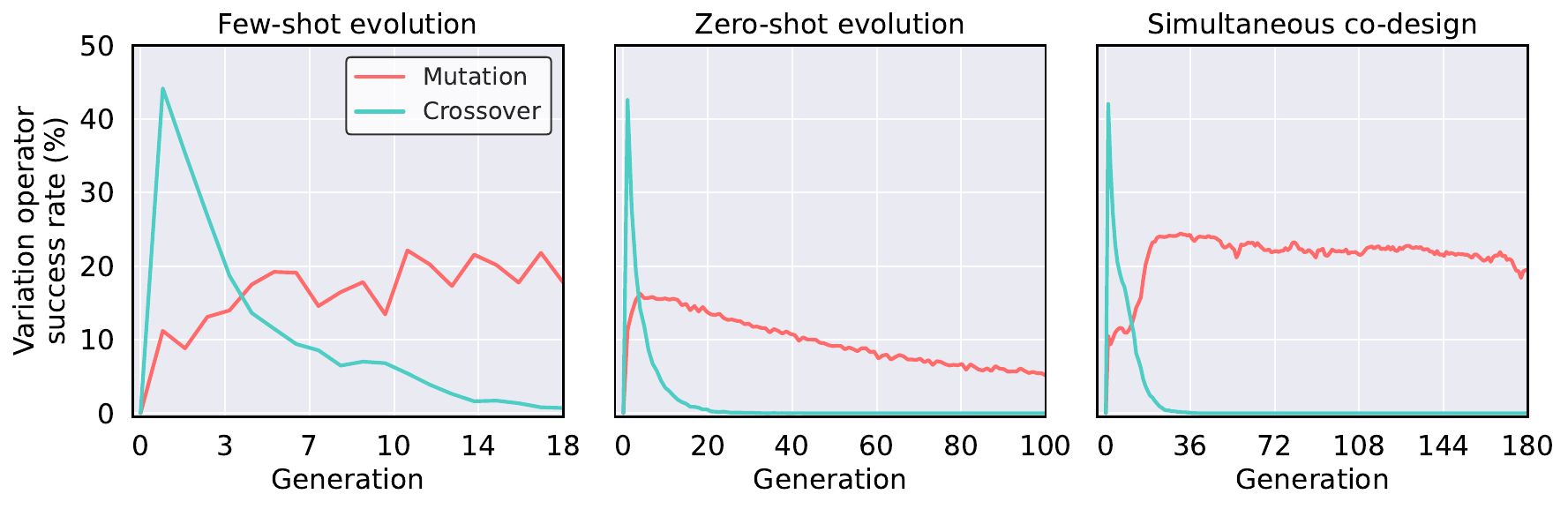}
    \vspace{-20pt}
    \caption{\textbf{Success of crossover vs.~mutation.}
    Early in evolution, random crossover is more successful than random mutation.
    But, after a few generations, mutations that more finely tune good designs were less likely to be deleterious than swapping large components between designs. 
    }
    \vspace{-8pt}
    \label{fig:appendix-variation-operator-stats}
\end{figure*}

\begin{figure}[b]
    \centering
    \includegraphics[width=0.5\textwidth]{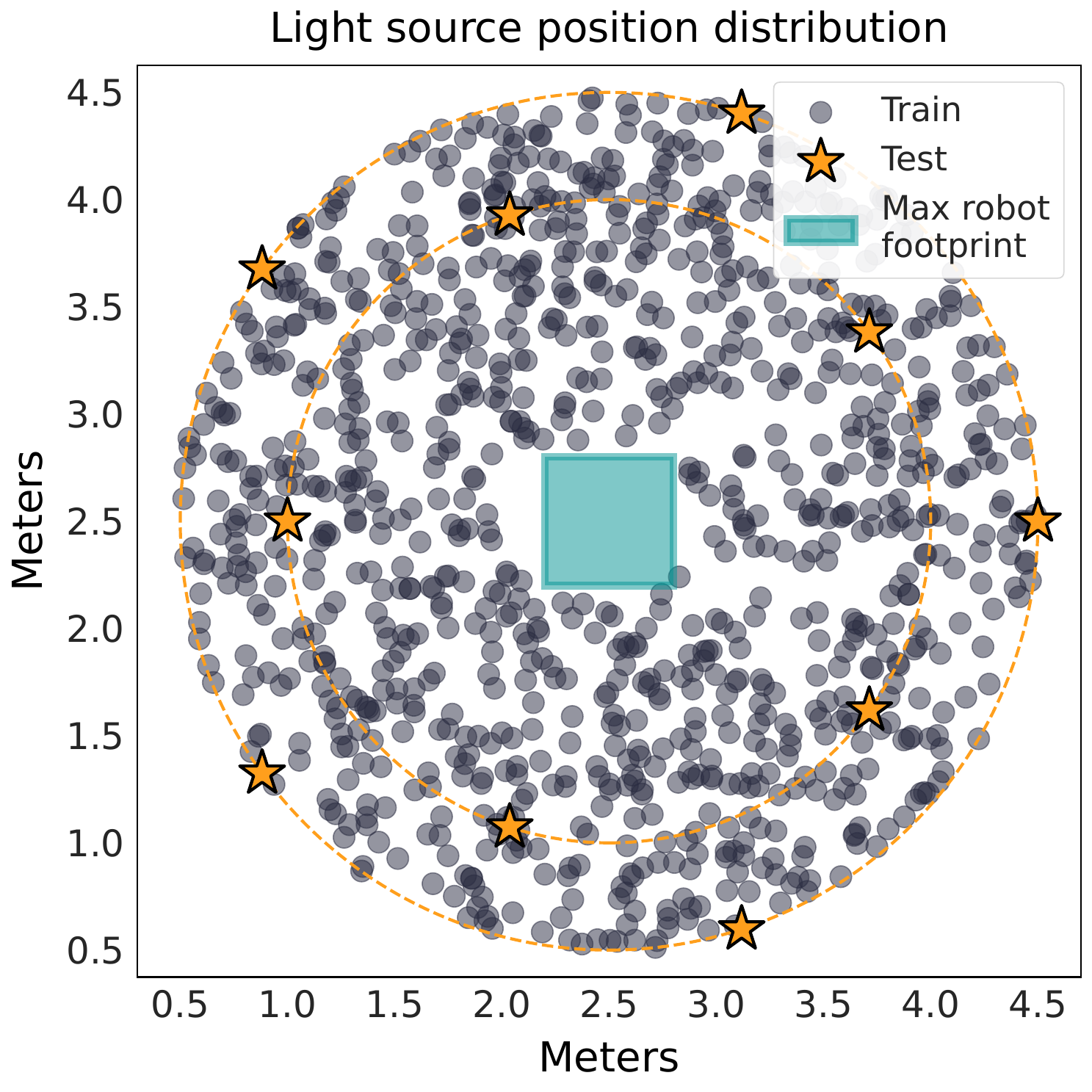}
    \vspace{-8pt}
    \caption{\textbf{Phototaxis training and testing.}
    During pretraining, simult.~co-design, and few-shot finetuning,
    training light source locations (gray circles) were sampled uniformly within 
    a circle centered on the robot's initial position (blue square).
    At every learning step,
    a batch of 8192 randomly positioned lights was sampled,
    and each was paired with a unique, random morphology and random terrain.
    Test light source locations (orange stars) 
    were identical across all methods for fair comparison.
    }
    \vspace{-8pt}
    \label{fig:methods-lightsource-dist}
\end{figure}

\begin{figure}[t]
    \centering
    \includegraphics[width=0.65\textwidth]{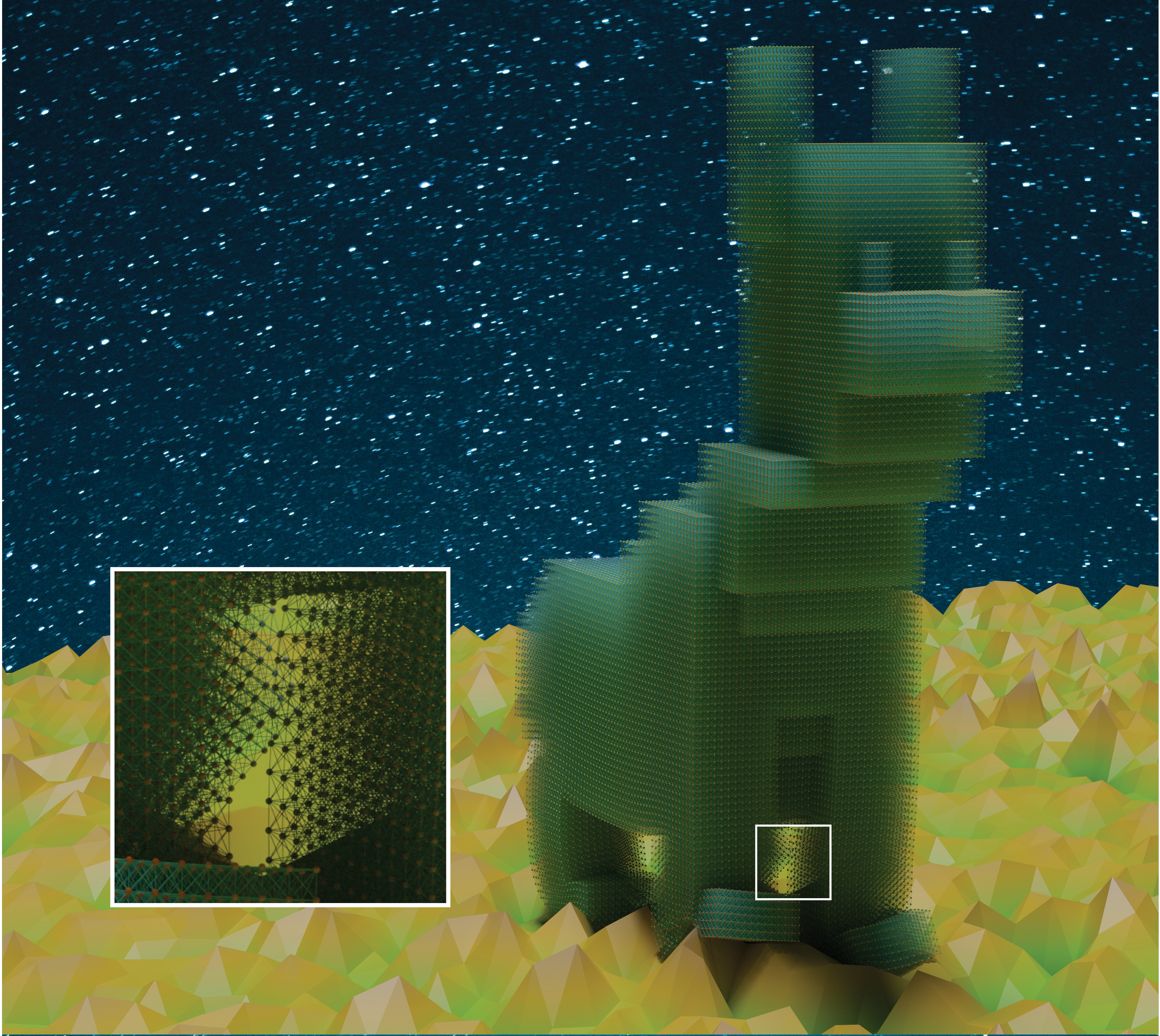}
    \caption{\textbf{Scaling morphology.}
    The embarrassingly parallel nature of the co-design pipeline
    allows the compute required to simulate 1024 robots with up to 1648 springs 
    (i.e.~a single GPU)
    to be redistributed 
    for a single robot with 1,115,157 springs.
    }
    \vspace{-10pt}
    \label{fig:results-dog}
\end{figure}

\end{document}